%% file: paper.tex
\documentclass[letterpaper, 10 pt, conference]{ieeeconf}  % Comment this line out if you need a4paper

\IEEEoverridecommandlockouts                              % This command is only needed if 
\usepackage{float}
\usepackage{times}
\usepackage{epsfig}
\usepackage{graphicx}
\usepackage{amsmath}
\usepackage{amssymb}
\usepackage{makecell}
\usepackage{multirow}
\usepackage{svg}
\usepackage{caption}
\DeclareMathOperator*{\argmax}{argmax}
\DeclareMathOperator*{\E}{\mathbb{E}}
\usepackage{tabularx,booktabs}
\newcolumntype{C}{>{\centering\arraybackslash}X} 
\usepackage{soul}

\usepackage{cite}
\usepackage[pagebackref=true,breaklinks=true,letterpaper=true,colorlinks,bookmarks=false]{hyperref}

\title{\LARGE \bf
AutoPhoto: Aesthetic Photo Capture using Reinforcement Learning
}

\author{Hadi AlZayer$^{1}$, Hubert Lin$^{1}$, and Kavita Bala$^{1}$% <-this % stops a space
\thanks{$^{1}$All of the authors are with the Department of Computer Science, Cornell University, Ithaca NY 14850, USA {\tt\small \{ha366,hl2247,kb97\}@cornell.edu}.}%
}

\begin{document}

\maketitle
\thispagestyle{empty}
\pagestyle{empty}
%% PAGE NUMBERS: COMMENT THIS OUT BEFORE SUBMISSION %%
% \thispagestyle{plain}
% \pagestyle{plain}
% \hubert{Comment out page numbers before submitting}
%%%%%%%%%%%%%%%%%%%%%%%%%%%

%%%%%%%%%%%%%%%%%%%%%%%%%%%%%%%%%%%%%%%%%%%%%%%%%%%%%%%%%%%%%%%%%%%%%%%%%%%%%%%%
\begin{abstract}
   The process of capturing a well-composed photo is difficult and it takes years of experience to master. We propose a novel pipeline for an autonomous agent to automatically capture an aesthetic photograph by navigating within a local region in a scene. Instead of classical optimization over heuristics such as the rule-of-thirds, we adopt a data-driven aesthetics estimator to assess photo quality. A reinforcement learning framework is used to optimize the model with respect to the learned aesthetics metric. We train our model in simulation with indoor scenes, and we demonstrate that our system can capture aesthetic photos in both simulation and real world environments on a ground robot. To our knowledge, this is the first system that can automatically explore an environment to capture an aesthetic photo with respect to a learned aesthetic estimator. \textit{Source code is at} \href{https://github.com/HadiZayer/AutoPhoto}{https://github.com/HadiZayer/AutoPhoto}
\end{abstract}

%%%%%%%%%%%%%%%%%%%%%%%%%%%%%%%%%%%%%%%%%%%%%%%%%%%%%%%%%%%%%%%%%%%%%%%%%%%%%%%%

\input{sections/intro}

\input{sections/related_work}

\input{sections/problem_setup}

\input{sections/rl_setup}

\input{sections/aesthetic_model}

\input{sections/implementation}

\input{sections/results}

\input{sections/discussion}

%\section*{APPENDIX}

\section*{ACKNOWLEDGMENT}

We thank Mark Campbell, Yutao Han, Jacopo Banfi, and Vikram Shree for assistance with the Jackal; Utkarsh Mall for providing an implementation of the user study framework; and Aaron Gokaslan for helpful discussions. This work was funded in part by NSF (CHS-1617861 and CHS-1513967) and NSERC (PGS-D 516803 2018).

\bibliographystyle{bibtex/IEEEtran}
\bibliography{paper}

\addtolength{\textheight}{-12cm} % This command serves to balance the column lengths

\end{document}

%% file: sections/intro.tex
\section{Introduction}

Cameras are now widely accessible to most people, but taking a well-composed photo is a difficult task that requires significant practice and experience. With advances in autonomous agents, there is an increasing interest in leveraging drones or robots to reduce human effort in various domains. For example, automatic camera planning can be used to capture sports events \cite{chen2016learning}, and drones can be used to create cinematographic videos \cite{bonatti2020autonomous, jiang2020example, huang2019learning}. Autonomous agents are also well-suited for use in remote, dangerous, or otherwise difficult-to-access locations (like caves, forests, or, in an extreme case, other planets like Mars). In real estate, marketing properties requires carefully composed photographs that showcase indoor architecture and layouts. The process of cataloguing different properties is time intensive for a human agent, and physical access to properties may be limited (due to, for example, the recent COVID-19 pandemic). Finally, an autonomous photography system can also be used to guide novice photographers towards better composed photos.

%Capturing aesthetic photos of indoor environments can also be a great asset for real estate, for example for individuals/agents interested in listing their properties on platforms like Zillow and AirBnB. 
%Such a system can also guide novice human photographers towards capturing an aesthetically pleasing photo with respect to a learned aesthetics function.
%\hadi{since initial views are not related -- demonstrate why we want to take an aesthetic view unrealted to init. Ex. exploring the scenes -- take a good shot every few meters. Reduce the time required}

Our goal is to build an autonomous system that can capture aesthetically pleasing photographs. Relying on heuristics is one way to compose aesthetic photographs. For example, the rule of thirds is a heuristic in which important objects of interest are aligned with imaginary lines that divide an image into thirds along horizontal and/or vertical axes. However, heuristics do not fully capture human aesthetics preferences. Indeed, recent work \cite{ven, vfn, personalized_aesthetics2017adobe, personalized_aesthetics2019,ava,chen2017quantitative} has focused on learning subjective preferences for aesthetics directly from humans. These aesthetics models can implicitly understand what makes well-composed photographs beautiful, but this understanding cannot be easily decomposed into explicit rules like the rule of thirds.

\begin{figure}[ht!]
  \centering
  %\vspace{-3cm}
  \includegraphics[width=1.0\linewidth]{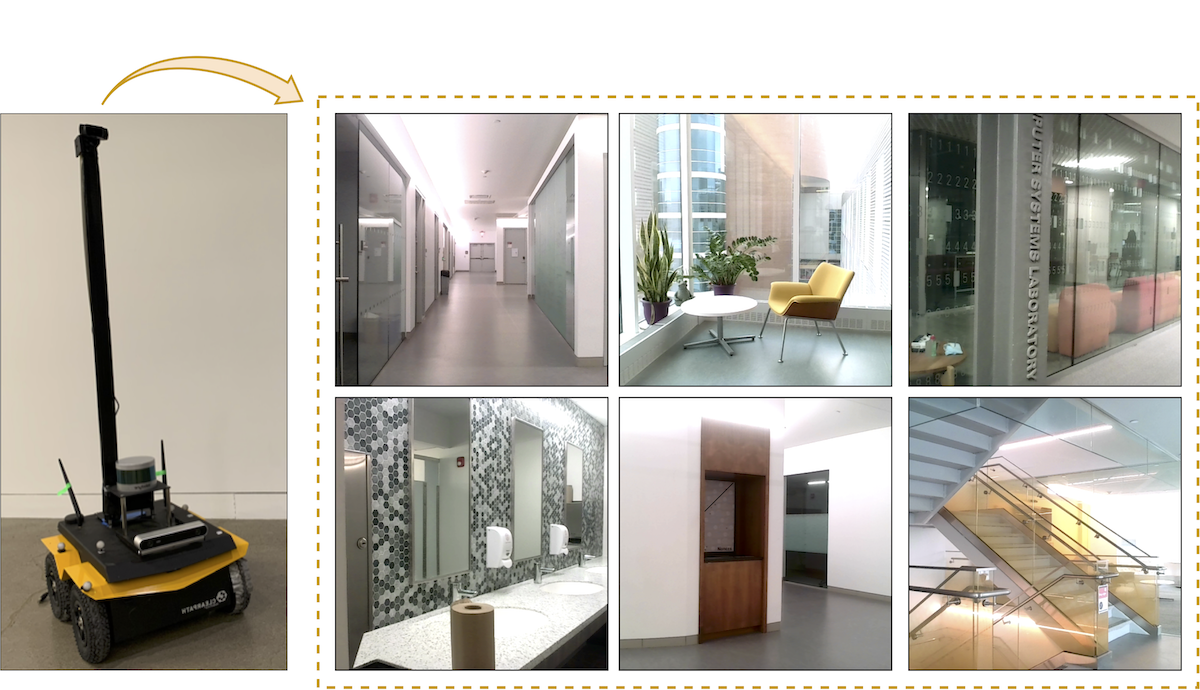}
    %\includegraphics[width=0.49\linewidth]{figures/state.pdf}
%   \includegraphics[width=0.24\linewidth]{figures/architecture.pdf}
%   \includegraphics[width=0.24\linewidth]{figures/architecture.pdf}
%   \includegraphics[width=0.24\linewidth]{figures/architecture.pdf}
  %\vspace{-15cm}
  \caption{\textbf{Photos Captured By AutoPhoto.} Left: The AutoPhoto system deployed on Clearpath Jackal robot. Right: Photos autonomously captured by AutoPhoto. See Fig. \ref{fig:real_results} for comparisons against initial environment views.}
  \label{fig:teaser}
\end{figure}

In this work, we present AutoPhoto, a system that sequentially takes actions to explore an environment, with the end goal of capturing an aesthetic photograph. Two important branches of work in autonomous aesthetic composition are (a) image cropping \cite{ven, vfn} and (b) drone cinematography \cite{bonatti2020autonomous, huang2019blearning}. Image cropping is a limited form of ``pseudo-photography'' with a fixed viewpoint and known environment (i.e., the original uncropped image is the environment from which the cropped photo should be captured). In this work, we are interested in a more general setting in which the photographer must explore an unknown environment via different viewpoints. Although autonomous cinematography is also concerned with varying viewpoints in unknown environments, it is constrained by tradeoffs between smooth motion planning, temporal constraints, and the final aesthetics of the film. Further, cinematography is not strictly concerned with aesthetic view capture; instead, it is focused on capturing events or telling a story \cite{mascelli1965five, gleicher2008re}. As a result, some existing work simplify aesthetic estimation by leveraging heuristics like shot templates or optimizing rule of thirds with respect to actors in a scene \cite{bonatti2020autonomous, visigrapp, photographer-robot, auto_composition}. In this paper, we are interested in optimizing image aesthetics with respect to data-driven aesthetics models to better capture human preferences.

A challenge with photo viewpoint optimization with respect to a
learned aesthetic function is that it is difficult to formulate
analytically (contrast this with the rule of thirds which can be
directly optimized given an object of interest, e.g.
\cite{visigrapp,auto_composition}). As such, we use reinforcement
learning to learn a controller that can navigate to views that are
aesthetic with respect to a learned aesthetic model. A second
challenge lies in the aesthetic function itself. Existing work in
learning image aesthetics has focused primarily on its application in cropping \cite{ven, vfn} where models are trained to compare crops
from within the same image. However, the task of photo capture
requires views from different parts of an environment to be compared,
and the aesthetics function should be robust to variations that may
naturally arise such as minimal camera translation, and a change in camera exposure. We propose an aesthetics model which is better suited for this task than existing cropping-based aesthetics models. Given this aesthetics model, we demonstrate that we can learn to navigate an environment to capture aesthetic photos. Our system is trained in simulation with realistic indoors reconstructed scenes using the Gibson dataset~\cite{gibson} with AI Habitat~\cite{habitat}. Our experiments demonstrate generalization to unseen scenes across simulation and real life.

This paper is organized as follows. First, we review related work in
Section \ref{sec:related}. We describe our problem formulation in
Section \ref{sec:problem_setup}. Then, we describe our reinforcement
learning pipeline in Section \ref{sec:rl_pipeline}. We propose an
improved aesthetics estimator in Section \ref{sec:aesthetic} suited
for the task of photo capture. We cover implementation details in Section \ref{sec:experiments}. Finally, we present the results for our AutoPhoto system in Section \ref{sec:results}, with quantitative evaluations in simulation and real life, including evaluation against human preferences.

% We demonstrate how the model performance translates to real life by guiding human users to take photos. 

% While aesthetics can be subjective, it is possible to learn a model to estimate the aesthetic value of a photo by learning to rank photos of different aesthetic value \cite{vfn, ven}. It is possible to learn an aesthetic estimator by learning to rank the images in a manner consistent with user ranking \cite{ven}, or rank the original image better than the same image corrupted in some manner\cite{vfn}. Using a learned aesthetic model as our objective can generalize to any type of photos that the aesthetic model was trained on. Even though there exist common heuristics like the rule-of-thirds, they are not sufficient to autonomously decide where to take the photo.

Our contributions are:
\begin{itemize}
%    \item A formulation of aesthetic photo capturing as a partially observable Markov decision process that can be optimized using reinforcement learning (RL). \hubert{remove? or merge with next pt somehow}
    \item A novel reinforcement learning pipeline for a generalized photo capture problem which includes (a) an unknown environment, (b) photographer movement and rotation, and (c) optimization against a learned aesthetics estimator that models human preferences better than heuristics.
    \item An aesthetics model that is consistent with human preferences across diverse viewpoints, and is robust to variances  encountered while taking photographs.
    \item Experimental validation that demonstrates successful navigation in unseen environments across both simulation and real life to capture aesthetically pleasing photos. We deploy the system autonomously on a Clearpath Jackal UGV.

\end{itemize}

\begin{figure*}[ht!]
  \centering
  %\vspace{-3cm}
  \includegraphics[width=1.0\linewidth]{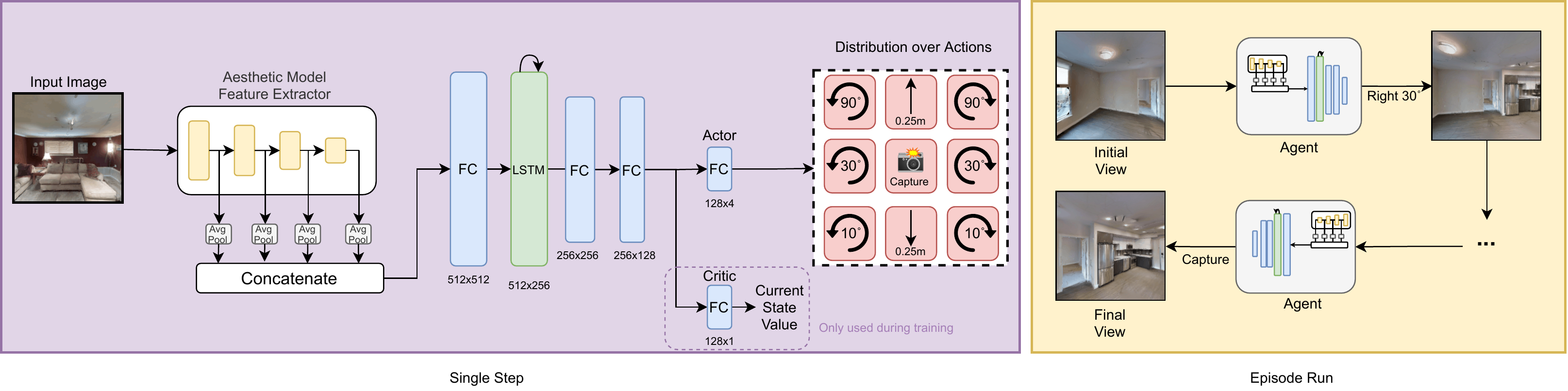}
    %\includegraphics[width=0.49\linewidth]{figures/state.pdf}
%   \includegraphics[width=0.24\linewidth]{figures/architecture.pdf}
%   \includegraphics[width=0.24\linewidth]{figures/architecture.pdf}
%   \includegraphics[width=0.24\linewidth]{figures/architecture.pdf}
  %\vspace{-15cm}
  \caption{\textbf{Illustration of the Pipeline and Runtime
Execution.} Left: AutoPhoto is composed of an aesthetics model that extracts features from the current view, a common MLP+LSTM backbone that processes these features, and two separate layers that parameterize the actor and critic. The actor selects an action to take and the critic estimates the current state value. We iteratively run multiple episodes to sample action
and state value pairs to optimize the model parameters. Right: During inference, the model takes a sequence
of movement actions until the agent (actor) selects the capture action.}
\vspace{-2mm}
  \label{fig:arch}
\end{figure*}

%% file: sections/related_work.tex
\section{Related Work \label{sec:related}}

\paragraph{Image Aesthetics}

%To be able to find aesthetically pleasing views, we need to quantify aesthetics to be able to optimize for a view with respect to its aesthetic value. Some existing work in autonomous composition use heuristics like the rule of thirds  \cite{visigrapp, photographer-robot, auto_composition} or template matching \cite{drone_pair, portraits} for automatic view selection. % talk about how they optimize for the objective. Also mention that they primarily focus on the drone planning problem

Understanding human judgements of image aesthetics has a long history in psychology and neuroscience, and photographers follow well-known rules to capture aesthetic images \cite{deng2017image}. For automatic view selection, such rules can be captured by heuristics like the rule of thirds and template matching \cite{bonatti2020autonomous, visigrapp, photographer-robot, auto_composition}. Computer vision models have also been developed to estimate the aesthetics of images \cite{deng2017image}. Modern aesthetics models are trained on large numbers of human judgements, and can generalize better to scenes where no clear heuristics can be applied. 
%For example, the rule of thirds requires a defined object of interest and template matching requires suitable templates for the current environment, but these requirements may not be met in general.
%Learning a model to estimate the aesthetics of an image can generalize better to scenes where there are no clear heuristics to follow. For example, a learned aesthetics model is useful when there are no defined objects of interest and the goal is to take an aesthetic image of the environment itself, or when the heuristics do not adequately define a target composition.  
Recent work learns to rank the aesthetics of pairs of images \cite{vfn, ven}. In \cite{vfn}, the model is trained to rank images against random crops of the same image. The assumption is that the composition of professional images are well-balanced while a random crop is less balanced. Alternatively, learning directly from human preferences is possible given datasets of aesthetic score judgements of crops or images \cite{ven, chen2017quantitative, ava}. %can be collected by asking humans to rank different crops \cite{ven, chen2017quantitative} or different images \cite{ava}.
%With an annotated dataset, aesthetics models can be trained to rank images according to human preferences. 
Aesthetic models can also learn personal preferences, where the scores can vary from user to user \cite{personalized_aesthetics2017adobe, personalized_aesthetics2019}.
%  However, this work is limited to taking portraits due to the problem structure as it uses body pose estimate to find the nearest-neighbor image. Our work can be generalized across image genres as we do not make restrictions on what type of photos we can take.

% In our work, we show that given an aesthetic value function, we can learn a model that takes actions to find a view that optimizes for the given aesthetic estimator. \\
% Also should discuss Markov Decision Processes and Reinforcement Learning here. One caveat is in our setting, it is a Partially Observed Markov Decision Process -- so some sort of exploration is necessary. Explain states, actions, and rewards. Then briefly explain PPO. Maybe this should be in the problem formulation? But also need relevant work in applying reinforcement learning in the related work section.

% \vspace{-4mm}
\paragraph{Automatic Photo Composition}
Existing work on automatic photo composition primarily focuses on image cropping, where the composition of photos are improved post-capture. Both common heuristics like the rule of thirds and visual balance \cite{auto_composition} and learned aesthetic scores \cite{vfn, ven} are used for automatic image cropping. 

To be able to use the rule of thirds and visual balance for image composition, one needs to specify an object of interest. \cite{auto_composition} extracts salient objects from the image, then automatically composes the photo by using the saliency mask to compute scores for the rule of thirds, visual balance, and diagonal dominance. Creatism \cite{creatism} mimics the pipeline of a professional landscape photographer by cropping and post-processing panoramas from Google Street View. % Other work \cite{portraits} uses nearest neighbor search to suggest portrait photo composition ideas by querying a large dataset of high quality portraits to retrieve photos similar to the current setting.

On the other hand, a learned aesthetics model is able to capture more nuanced composition rules that are not captured by explicit heuristics. To compose photos using a learned aesthetic model, \cite{vfn} uses a sliding window with various aspect ratios to select the crop with the highest aesthetic score. Since a sliding window approach is computationally expensive, \cite{a2rl} uses a reinforcement learning model to sequentially adjust the crop window.
% While improving photo composition through cropping is an important post-processing step, an important problem is capturing aesthetic photos in the first place -- this is the task that we consider in this paper. Unlike cropping, aesthetic photo capture involves navigating and exploring 3D scenes to find an aesthetic view. The act of movement allows a photographer to handle occlusion and find view compositions that may not be visible from their initial vantage point.  \hadi{comparison between cropping and photo capture can be abridged}
%Applications of aesthetics to image cropping (cite some of the same papers). Similar to us bc we want to find aesthetic view. Different because cropping assumes environment is known (env. is full original uncropped image).

% \vspace{-4mm}
\paragraph{Drone Cinematography}
% Previous work on drone photography covers tracking actors and selecting angle
Autonomous cinematography and camera planning have received increasing attention in recent years \cite{bonatti2020autonomous, jiang2020example, drones_cmu, huang2019learning, chen2016learning}. One way to compose videos is to optimize paths between key frames. These key frames can be predefined by a user \cite{tog_keyframes, airways} or sampled intelligently from a set of template shot types \cite{bonatti2020autonomous, drones_cmu}. \cite{drone_pair} automatically records videos of two subjects by matching the view with predefined templates for scenes with two actors.  Instead of defining key frames, imitation learning \cite{hussein2017imitation} can be used to learn camera trajectories directly from films. \cite{huang2019blearning, huang2019learning} apply imitation learning to learn from human-created films while \cite{jiang2020example} focuses on learning from examples for computer-animated cinematography.
Recent work has also explored semantic control over the emotions evoked by clips by learning from annotated video clips \cite{bonatti2020batteries}.
 %There are also commercial drones (e.g., Skydio) that allow for autonomous cinematography by tracking a user, even during high-speed activities like skiing and mountain biking.

A key challenge in cinematography lies in temporal consistency -- the aesthetics of the final video depends on frames captured in real-time. This challenge limits the ease of integrating learned aesthetics models into drone cinematography. As such, a common property of the many autonomous cinematography works is that the captured video focuses on tracking human subjects \cite{huang2019blearning, huang2019learning, drone_pair, chen2016learning, drones_cmu, bonatti2020autonomous}. Commercial drones (such as Skydio) also focus on autonomous cinematography for a single user by filming them during activities like skiing or mountain biking. By focusing on human subjects, heuristics for framing human actors can be readily applied instead of optimizing for aesthetics in a general setting.

% \hadi{cite more autonomous cinematography papers here}
%-- it can be unclear how to optimize paths over learned aesthetics while path optimization over heuristics is relatively simpler. 
% In contrast, our task of photo capture does not require the full sequence of views to be aesthetic; rather, only the final view needs to be well-composed, and the photographer is free to explore the environment prior to capturing the final photo.

% similar to our problem: navigate 3D environment, find aesthetic shot for videos. focuses on heuristics. similar to us because we want to navigate to a view in a 3d environment. Different bc we are exploring instead of focusing on optimizing trajectory, and also bc we are trying to learn structure from the aesthetic models 

%% file: sections/problem_setup.tex
\section{Problem Setup and Pipeline \label{sec:problem_setup}}
%%%%%%%%%%%%%%

Our objective is to autonomously explore a local region within an environment to capture an aesthetic photo.

To take a photo, a photographer assesses a view through the camera viewport, and then iteratively adjusts the composition through movement and rotation of the camera. When a balanced composition is achieved, the photographer captures the photo. Our goal is to create an agent that can achieve this functionality. This process can be described as a series of actions: (a) observe the current view, (b) move the camera if the view can be improved and repeat from (a), or (c) capture the photo if the view is sufficiently well-composed and terminate.
% The person observes the view through the camera view port, evaluates the view quality, if there is a room of improvement then they move the camera and restart the loop, but if the view achieves a balanced composition then the user captures the photo and terminate.
To judge the quality of the image, we assume that we have oracle
access to an aesthetic value function $\phi$ that maps an image to its
aesthetic score. Our objective is to optimize the
aesthetic score of the final captured image, so that the aesthetic score of the final image achieves some threshold. To generalize
to a variety of scenes, we introduce an aesthetic score threshold that
is scene dependent and invariant to the scene size. This threshold is
based on the estimate of the mean of aesthetic scores and score
variations in the local region to make the threshold independent from the scene size.
We discuss how an aesthetics model can be learned (Section
\ref{sec:aesthetic}), and how it can be used in practice to set the
threshold (Section \ref{sec:rl_training}).

We formulate the problem as a partially observable Markov decision process (POMDP) -- the environment is not fully observable as the agent only has access to partial views of the environment
through the camera viewport. The actions are camera movements and
photo capture. %, and the rewards are defined by the aesthetic value of the captured image. %To solve a POMDP with reinforcement learning, the learned policy will include a memory state \cite{wierstra2010recurrent}. 
In Section \ref{sec:rl_pipeline}, we discuss this formulation in detail. %, and describe how reinforcement learning (RL) can be used to solve the problem. 

In Fig. \ref{fig:arch} we demonstrate our RL-based pipeline. On the
left we show the architecture details to perform a single step, and on
the right we show the action sequence taken by the agent at runtime.
To train our model, we run multiple episodes and compute the reward
function after every step to sample state-action reward pairs, then we
train the Actor based on the current Critic state value estimates, and
train the Critic using the collected state-action reward pairs.
% \hadi{do we want to move the training sentence to 4.3? Also this is
% just the general idea for how actor-critic algorithm works, but PPO
% involves reward clipping and learning rate based on the amount of
% change between the old policy and new policy. Should we include more
% details? less?} \kb{I think this level of detail is good. And then
% leave a forward pointer that more details on the policy will be given
% in Section XXX.} 

% \kb{Then given roadmap of the rest of the paper. In sec 4 we talk
% about XX, In Sec 5 we talk about XXx... } \hadi{we already have a roadmap throughout the section I feel}

% \kbinline{In Figure XXX we demonstrate our RL-based pipeline. On the
% left we show the setup for training; on the right we show the action
% sequence taken by the agent at runtime.} \kb{Now describe the new
% pipeline Figure 1 and describe how it is trained, and how it is used
% in runtime.}

%% file: sections/rl_setup.tex
\section{RL for Aesthetic Image Capture \label{sec:rl_pipeline}}
% We now describe how to use the POMDP formulation to solve our problem. 
We describe the objective, state space, action space, and reward functions such that maximizing the objective corresponds to capturing an aesthetically pleasing photo.
% In Section \ref{sec:state} we describe the space of states and
% actions, and in Section \ref{sec:rewards} we define a reward function,
% such that maximizing Eq. \ref{eq:pomdp} would correspond to
% capturing photos with a high aesthetic score.
%an aesthetic score that achieves the aesthetic
%threshold $\tau_{aes}$ defined in Section \ref{sec:rl_training}. 
We will detail our architecture and implementation in Section \ref{sec:experiments}.

%\noindent{\bf Problem formulation.} 
The objective is to find a policy that maximizes the expected sum of discounted rewards. Specifically, denote $\pi$ to be a policy that maps to a distribution over actions given a state $s_t$, and let $\rho^\pi$ be the probability distribution over possible reward realizations $R = r_0,r_1,...,r_{T-1}$ based on the state-action trajectories given $\pi$. Formally, the objective is to find an optimal policy $\pi^*$ that satisfies 
\begin{align}
\label{eq:pomdp}
    \pi^* = \argmax_\pi \E\limits_{R \sim \rho^\pi}[\sum\limits_{t=0}^{T-1} \gamma^t r_t]
\end{align}
where $\gamma$ is a discount factor in $[0,1]$ that reduces the weight
of rewards far in the future.
%\hadi{The correct objective is
% "Expected" sum of discounted rewards. Should we switch it to that?
% would involve introducing some additional notations like policy $\pi$
% and states $s_t$ and actions $a_t$ and making $r_t$ to be a function}
% \kb{Yes, do what is the correct objective, assuming it is what you
% implemented, and describe what you need. The reviewer will be
% sophisticated.}

% \subsection{State and Action Spaces \label{sec:state}}
% \vspace{-4mm}
\paragraph{State and Action Space} %\hadi{add discussion to say why adding the fine and large turns are useful compared to the default actions}
We set the state to be the current camera view and an
LSTM memory cell. Memory enables RL-based solutions for a POMDP \cite{wierstra2010recurrent} by allowing the agent to utilize information from its exploration history to make decisions.
A memoryless agent makes decisions based only on the current view, which can lead to sub-optimal decision making as
it might re-explore regions with low aesthetics, and potentially lead to a non-terminating loop.
%despite having
%seen it earlier in the episode. In an extreme case, this could lead to
%a non-terminating loop.

Our action space consists of the following actions: forward and backward for $0.25$m, and turning right and left by $10^\circ$, $30^\circ$, and $90^\circ$, and finally the CAPTURE action to take the photo and terminate the episode. A composition of these actions allows the agent to navigate to any position and orientation in the search space, modulo the regions which lie between the discretizations of the actions defined above. The values of $0.25$m translation and $30^\circ$ rotations are the default in \cite{habitat}. We introduced more turning angles to allow better control between fine turns to adjust the view composition and large turns to quickly explore the scene.
% \begin{figure}[!ht]
%   \centering
%   %\vspace{-3cm}
%  \includegraphics[width=\linewidth]{figures/state.pdf}
%   \caption{state and actions}
%   \label{fig:arch}
% \end{figure}

% \subsection{Objective and Rewards\label{sec:rewards}}
% \vspace{-4mm}
\paragraph{Objective and Rewards}
The objective is to capture a final view $s_T$ that has a high aesthetic score according to an aesthetic estimator $\phi$ (i.e., $\phi(s_T) > \tau_{aes}$). $\tau_{aes}$ is set according to the aesthetics of the local region (see Section \ref{sec:rl_training}).
%to capture a photo corresponding to the view at termination $s_T$ with a high aesthetics score (i.e., $\phi(s_T) > \tau_{aes}$). $\tau_{aes}$ is defined adaptively based on each scene -- refer to Section \ref{sec:rl_training} for details. \hadi{use $s_T$ here too? will also need to move the clarification on $s_T$ after the reward func. if change it}
% The aesthetic score that we require the captured image to achieve is based the mean $\mu$ and standard deviation $\sigma$ of the k nearest neighbor views to the initial camera position of the $N$ randomly sampled views across the scene. The parameters k and $N$ can vary across the scenes but they should be large enough to capture the region of interest to the user and provide an accurate estimate of the scores across the scene, respectively.
We set the reward for the CAPTURE action to be: \begin{align}
    r(s_T, \text{CAPTURE}) = \begin{cases} +1 & \phi(s_T) > \tau_{aes} \\
    -1 & \text{otherwise}\end{cases}
\end{align}
However, a final reward is not sufficient to train the RL agent on its
own since it is sparse, and does not take into consideration the
number of steps taken to capture the photo. It is important to learn
an efficient policy which takes as few steps as possible. If
efficiency is not of concern, one could run an exhaustive grid search
to find a view that maximizes the aesthetic score, but this is
impractical in general. For movement actions where $a \neq$ CAPTURE,
we define the step reward as:
\begin{align}
    r(s_t, a) = \phi(s_{t+1}) - \phi(s_{t}) + 0.1 \Gamma(\zeta) - \beta t \label{eq:non-terminal-reward}
\end{align}
$(\phi(s_{t+1}) - \phi(s_{t}))$ is the score difference between
the current and next view to encourage the agent to move towards
regions with increasing aesthetic score. $\Gamma(\zeta)$ is an
exponentially decaying exploration reward where $\zeta$ is the number
of steps since training has started (in contrast, $t$ is the number of steps in the current episode). The exploration reward encourages
the model to explore and avoid terminating during the early stages of training; this is similar to intrinsic curiosity rewards as in
\cite{curiosity}. Finally, $\beta$ is a time step penalty to
discourage the model from taking too many steps. In our experiments,
the exploration reward $\Gamma(\zeta)$ is $0.9999^{\zeta}$ and we set the time step penalty $\beta$ to be 0.005. As we show in our ablation studies in section \ref{sec:ablation}, the step reward for non-terminal actions is critical for good performance.

%% file: sections/aesthetic_model.tex
\section{Aesthetics Model \label{sec:aesthetic}}

We now describe the aesthetics model we use to model human aesthetics preferences. This model will be used to generate rewards for the RL agent during training. 

% \vspace{-4mm}
\paragraph{Ranking Views} The task of photo capture requires estimating the aesthetics of images across different viewpoints. Existing aesthetics models are typically trained for cropping, and do not learn to rank images with different viewpoints \cite{ven, vfn}.  The Aesthetic Visual Analysis (AVA) dataset \cite{ava} contains scores for images of different content and view points. However, existing work \cite{ven} has shown that models trained solely on AVA struggle to perform well on cropping benchmark datasets such as Comparative Photo Composition (CPC), a dataset that contains rankings of different crops of each image. This suggests that AVA and CPC may contain complementary information about human aesthetics preferences, so we use both datasets. For AVA, photos are categorized into different genres, such as landscape and portrait. Because it is difficult to meaningfully rank images from different genres, the model is trained to rank pairs of images from the same genre. We use the standard pairwise ranking loss used in \cite{vfn, ven}. Let $s_1$ and $s_2$ be two images where $s_1$ should have a higher aesthetics score than $s_2$. The loss is:

\begin{align}
    \ell_{rank}(s_1, s_2) = \max(0, \phi(s_2)-\phi(s_1)+1)
\end{align}

% \vspace{-4mm}
\paragraph{Improving Robustness} We also consider losses to increase
the robustness of the model to camera translation and exposure. 
%An aesthetics model which is not robust to small translations can cause rapid score oscillations within a scene and result in instability when training the RL agent. 
Small translations are common as noise in camera movements is inevitable during real world deployment, and images should not be scored differently based on very small translations. To train the model to generate similar scores for similar images, we minimally crop images and minimize the Mean Square Error (MSE) between the score of the original image and the minimally cropped image. It is also useful for the model to rank well-exposed images better than over-/under-exposed images. Camera exposure can vary as lighting changes when a photographer navigates an environment. Since CPC and AVA only include well-exposed images, we introduce over-/under-exposed images by increasing/decreasing the brightness of the images. For an image $s$, the robustness loss is:
\begin{align}
    \ell_{robust}(s) &= \lambda_{sim}\ell_{sim}(s) + \lambda_{expo}\ell_{expo}(s) \\
    \nonumber \text{where } & \ell_{sim}(s) = \frac{1}{2}(\phi(s) - \phi(s_{\text{min crop}}))^2 \\
    \nonumber & \ell_{expo}(s) = \ell_{rank}(s, s_{\text{poorly exposed}})
\end{align} 

\noindent Our full loss function for the aesthetic model is:

\begin{align}
    \mathcal{L}_{aes}(s_1, s_2) = \lambda \ell_{rank}(s_1, s_2) + (1-\lambda) \ell_{robust}(s) 
\end{align} 
$s$ is selected from $\{s_1, s_2\}$ uniformly, and we set $\lambda = 0.6$, $\lambda_{sim} = 0.875$, and $\lambda_{expo} = 0.125$.

%% file: sections/implementation.tex
\section{Implementation \label{sec:experiments}}
% also discuss image sampling in each experiment

% Our system is trained in simulation and evaluated in both simulation and real life.

In this section, we describe the implementation of our system. Training in simulation is necessary since the model has to interact
with the environment for a large number of steps. There are different simulations that can be used such as AirSim \cite{airsim} and AI Habitat \cite{habitat}. We choose AI Habitat in our experiments due to its high frame rate and support for realistic indoor datasets like Gibson \cite{gibson} and Replica \cite{replica}.

In Section \ref{sec:aes_arch}, we describe the aesthetics model. In Section \ref{sec:rl_training} we describe the actor-critic RL model. 
%and its training details. 
%training process on the Gibson dataset.

\subsection{Aesthetic Model Implementation \label{sec:aes_arch}}

%\hubert{MERGE THIS INTO SECTION 5. SECTION 5: EXPLICITLY SPELL OUT HOW THE LOSSES ARE USED.}

\paragraph{Architecture} We use ResNet18 with a single scalar aesthetic score output. Modern CNNs are
inherently sensitive to small pixel translations, so we adopt an antialiasing solution \cite{invariant_cnn} by adding a blur layer during max pooling to increase robustness to small translations. 
% \hadi{is it clear that the paper is the source of the proposed solution?}
% \hubert{yes}

% \vspace{-4mm}
\paragraph{Training Details}

%For each sample in the training batch, we choose the training task to be aesthetic ranking with probability 60\%, minimal cropping with probability 35\%, or ranking well-exposed images against under-/over- exposed images with probability 5\%. 
Each image batch is drawn from CPC or AVA with equal probability. 
%We consider images from AVA with the tags cityscape, architecture, landscape, animal, and floral. 
Minimally cropped images are created by randomly cropping images between 1 and 5 pixels for each side. Over-/under-exposure is generated by multiplying brightness by 4 and 0.5 respectively. The model is trained for 210,000 iterations with batch size 32. 

% \subsection{RL Model Architecture \label{sec:rl_arch}}
\subsection{RL Model Implementation \label{sec:rl_training}}
% \hadi{should this section be combined with the Model Training subsection?}

\paragraph{Architecture} We use an actor-critic setup \cite{actor-critic} for the RL agent, illustrated in Fig. \ref{fig:arch}. The aesthetic model is used as a feature extractor for the camera view. The camera view features are computed by average pooling the output of each of the four residual blocks and concatenating them together.
%To extract the state features from a camera view, we use the trained feature extractor from the aesthetic model. We extract the features from the ResNet18 aesthetic model by average pooling the output of each of the four residual blocks and then concatenating them together. 
These features are given to an MLP with an LSTM layer which serves as a common backbone for the actor and critic layers. The output from the MLP+LSTM backbone is a combination of the current view and the memory state in the LSTM, and thus forms a representation of the current state. We use one classification layer for the actor to output a distribution over actions, and one layer for the critic to estimate the current state value. To optimize the architecture parameters, we use PPO \cite{ppo} using the stable-baselines implementation \cite{stable-baselines} with default hyper parameters.

% use rotation and 3D

% \kb{You might want to create a Section 6 that is about implementation
% and baselines etc. And put all of 6.1 and 5.2 and 5.3 there. Then
% Section 7 is all about results only and includes your current 4.4
% (baseline policies), 6.2, 6.3, 6.4.}

%(\kb{Once you move this to Section 6 (new), you
% will need to set the stage for why you are doing this subsection.})

% \begin{figure}[ht!]
%   \centering
% %   \vspace{-3mm}
%   \includegraphics[width=\linewidth]{figures/map_visualization_triple_v2.pdf}
%   \caption{\textbf{Top-down maps of scenes from Gibson.} Visualizations of the random samples collected and the samples that are used to compute the threshold (yellow) for an initial position (blue). \hadi{can be removed}}
%   \label{fig:map}
% \end{figure}

% \noindent{\bf Local region for aesthetic scoring.} 

\paragraph{Aesthetic Score Threshold}

The terminal reward depends on an aesthetic score threshold $\tau_{aes}$ that captured images should overcome. This threshold is set adaptively based on local regions within each scene, as aesthetics can vary across scenes and across sub-regions within a scene.

%To train our model, we need to define a threshold for an image aesthetic score to overcome to be considered aesthetically pleasing. The threshold must be scene dependent, since there is a natural variation of aesthetics across scenes (inside a palace as opposed to a small studio); and even within a scene, especially in large scenes with diverse sub-regions. Thus, to avoid dependence on scene size, the aesthetic threshold should be region-based within each scene. 

We define a local region in an environment by a set of points near
the agent's starting location. First, assume we have uniformly sampled some
set of N points (and their corresponding views) across the
environment. For any starting location, the K nearest neighbors define
a local region. The physical neighborhood defined by the KNN views
depends on the density of the N sampled points. The target aesthetics threshold for a local region is
set by considering the scores of the KNN views to the starting camera
position. Specifically, we use the mean aesthetic score of the KNN
views $\mu$ and the standard deviation of their scores $\sigma$ and
set the threshold to be:
\begin{align}
    \tau_{aes} = \mu + \sigma
\end{align}
The objective is to capture an image $s_T$ such that $\phi(s_T) >
\tau_{aes}$. This corresponds to the top 16\% views in the local region assuming the scores follow a normal distribution.

% We also \kb{delete consider? minimize} the distance from the starting location as
% a soft objective, which is discussed in Section
% \ref{sec:rl_pipeline}. 
% \vspace{-4mm}
\paragraph{Training Details}
We train the model using realistic reconstructions of indoor scenes
from Gibson \cite{gibson} with Habitat to run the simulation. We use a
subset of the Gibson dataset that was filtered by \cite{habitat} to
include only high quality reconstructions, and we remove any scenes
that include reconstruction artifacts that affect the scene
aesthetics. We split the subset of the Gibson dataset into 61 environments for training and 20 environments for evaluation. For each scene, we sample 2,000 random views, and we compute the aesthetic threshold using nearest 100 samples to the position of the initial camera. On every run, we sample a random navigable position and random orientation to set for the initial camera state and re-sample if the score of the initial view is too low (more than a standard deviation below the mean of the \emph{entire} scene). This is because a low-score initialization is likely in a region with poor aesthetics, and the model is unlikely to learn useful policies from such regions. Initialization re-sampling is not done during evaluation.
%We only remove low-score initialization during training but we keep it at test time. 
We train the model for 1.5 million steps, using a batch size of 8, and change the associated scene of each element in the batch after every 250 episodes to minimize the overhead of switching between simulated scenes.

%% file: sections/results.tex
\section{Results \label{sec:results}}

% In sections \ref{sec:eval_gibson} and \ref{sec:eval_matterport}, we discuss the evaluation results on Gibson and Matterport3D. Our results demonstrate generalization to unseen scenes from Gibson and cross-dataset generalization with Matterport3D. In Section \ref{sec:real_life} we show that our model can generalize to real life settings. We observe interesting emergent behavior that the model was not explicitly trained for, such as moving through doors, or using rotation judiciously to quickly explore surroundings. This behavior extends to our real-life experiments as well. Finally, we quantify the aesthetics of captured photos through a human study. \\

\emph{In the \href{https://youtu.be/RV830dZpQ-E}{video}, we include additional visualizations of initial and final view pairs, and clips of agent behaviors.}

In this section, we evaluate both our aesthetics model and the AutoPhoto system that is trained with our aesthetics model. In Section \ref{sec:aes_eval}, we compare the performance of our aesthetics model to an existing model \cite{ven}. In Section \ref{sec:baselines}, we describe the baseline policies that we compare AutoPhoto against. In Sections \ref{sec:eval_sim} and \ref{sec:real_life}, we evaluate the behavior of AutoPhoto on unseen environments in simulation. In Section \ref{sec:real_life}, we evaluate AutoPhoto in real life with human judgements. Finally, we include ablation studies to verify AutoPhoto design decisions in Section \ref{sec:ablation}.

% \kb{We evaluate the performance of our algorithm quantitatively in Table
% \ref{table:eval_threshold} and qualitatively with sample results in
% Figure \ref{fig:sim_results}.} \hadi{are references to the table and figures necessary here?} \\

\begin{table}[t]
    \centering
    \small
%       \caption{\textbf{Aesthetics Model Performance.} Ranking accuracy on CPC with ranking crops of the same
% images, and AVA with ranking different images of the same genre. Our
% model performs similarly well to Wei \emph{et al}. on a cropping dataset (CPC)
% but outperforms it significantly when ranking images of different
% viewpoints (AVA). Our model is also significantly more stable with
% respect to minimal cropping, and correctly ranks well-exposed images better than over- and
% under-exposed images.}
 \caption{\textbf{Aesthetics Model Performance.} Compared to \cite{ven}, our model performs similarly well in assessing aesthetics of image crops (CPC). However, our model performs better on cross-view ranking (AVA), ranks under-/over-exposed images below well-exposed images more often, and assigns more similar scores to nearly identical images. The latter properties are important for assessing aesthetics of different viewpoints under realistic conditions.}
    \begin{tabularx}{0.968\linewidth}{|c|c|c|c|c|} 
    \hline
    \makecell[c]{\emph{Task}} & \makecell[c]{Aesthetics \\ (CPC)} & \makecell[c]{Aesthetics \\ (AVA)} & \makecell[c]{Exposure} & \makecell[c]{Minimal \\ Cropping \\ (MSE)} \\
    \hline
    \makecell[l]{VEN} \cite{ven} & \makecell[c]{\textbf{75.8}} & \makecell[c]{62.2} &  \makecell[c]{81.0} & \makecell[c]{0.29} \\ 
    \makecell[l]{\bf Ours} & \makecell[c]{72.2} & \makecell[c]{\textbf{84.4}} & \makecell[c]{\textbf{99.7}} & \makecell[c]{\textbf{0.03}}\\ 
    \hline
    \end{tabularx}
  
    \label{table:aesthetic}
\end{table}

\subsection{Aesthetic Model Evaluation \label{sec:aes_eval}}

% \kb{This first sentence does not make clear we are measuring
% robustness to perturbations and small changes. Remind them again that
% is the goal.}
%We compare the aesthetic ranking performance and robustness of our aesthetics model against the aesthetics model from \cite{ven}. 
We evaluate the accuracy of rankings crops from CPC, as well as accuracy of rankings photos of different views (but the same genre) from AVA in Table \ref{table:aesthetic}. We also include translation robustness results as measured by MSE of scores between images and their minimally-cropped counterparts, and evaluate ranking accuracy with respect to under- or over-exposure. For reference, we compare our model to a state-of-the-art aesthetics model  \cite{ven}. With respect to ranking crops from the CPC dataset, our model performs a bit lower than \cite{ven}. However, the tradeoff is that our model performs better on cross-view ranking on AVA, and is far more robust to minimal translation and unflattering camera exposure.

\subsection{Baseline Policies for Photo Capture\label{sec:baselines}}
% Existing work in autonomous composition during the capture phase
% either focuses on video \cite{drones_cmu, drone_pair}, or uses the rule-of-thirds \cite{visigrapp} which requires user-defined objects of interest. Since our problem focuses on photo composition in environments without predefined objects of interest, it is difficult to compare against this work directly.

 We describe several baseline policies inspired by existing work. As discussed, existing work in autonomous composition during the capture phase cannot be directly applied as they focus primarily on video, and assume predefined objects of interest. For a fair comparison, we limit the total number of steps to 16 steps for the Key Frame Selection and Greedy policies to match the median number of steps taken by our method, as an unlimited number of steps would trivially achieve high accuracy. The policies are:

%To this end, we propose modified policies inspired by existing work to apply them in our setting. We implement a method based on the rule-of-thirds that explores the environment and navigate until a salient object is positioned in a manner that satisfies the rule-of-thirds. We also propose a stronger baseline that uniformly explores the environment and then captures the best scoring view found during the exploration phase, as well a greedy policy that  searches for a locally optimal view. \hadi{Can remove the preceding discussion because it repeats the descriptions we have later} For a fair comparison, we limit the total number of steps to 16 steps for the baselines to match the median number of steps taken by our method, as running for an unlimited number of steps would trivially achieve high accuracy.

% Instead, we focus on baselines that (a) are applicable to photo composition and (b) do not burden a user with additional input in the form of defining objects of interest in the environment.

% To this end, we propose a baseline policy \hadi{relies on key frame selection based on exploration history} inspired by post-capture image cropping models, in which well-composed image crops are chosen in a user-agnostic manner \cite{ven, vfn}. We also compare to a naive policy that acts randomly.

% \vspace{-4mm}
\paragraph{Random} uniformly samples actions.

% \vspace{-4mm}
% \paragraph{Cropping Based Policy} is inspired by cropping-based composition \cite{ven, vfn}. We design a policy that uses different crops from the current view to decide on an action to execute. Specifically, we consider the aesthetic scores of the crops corresponding to the leftmost two-thirds, center two-thirds, rightmost two-thirds of the current view. If the score of the full current view is higher than the scores of all the crops, then we select the CAPTURE action. Otherwise we select the action corresponding to the crop with the maximum score, where the actions corresponding to left, center, and right crops are TURN LEFT, FORWARD, and TURN RIGHT respectively.

\paragraph{Rule of Thirds} aligns an object of interest on the lower-left or lower-right third of the image. Since we do not have a predefined object, at each time step we compute the salient objects of the scene (similar to \cite{auto_composition, visigrapp}) using saliency detection network BASNet \cite{basnet}. If a salient object satisfies the rule of thirds, then CAPTURE is selected. Otherwise the agent takes a small turn, or moves to adjust the salient object position towards the lower-left or lower-right third of the frame. If no salient object is found, the agent takes a large turn to explore the environment as it is likely that the camera is facing a wall or featureless scene. 

\paragraph{Imitation Learning} learns actions directly from demonstrations \cite{hussein2017imitation}. We adapt \cite{huang2019learning} to our setting, whose model is trained to predict camera actions for cinematography. We generate demonstrations by sampling paths that lead to a local aesthetic maxima. Only demonstrations where the captured image exceeds the threshold for the current region are considered. Note that memory is important for modelling actions conditioned on paths. Since our proposed model already contains LSTM memory, we utilize the same architecture for this policy (except no critic branch). The model is trained on 17K demonstrations from Gibson with Adam, initial learning rate 1e-4, and exponential learning rate decay $\gamma$=0.95. We found 50 epochs to be sufficient, with 500+ epochs yielding no further improvements.

%uses a classifier to predict an action given an input state, which has been previously applied to cinematography \cite{jiang2020example, huang2019learning}. In imitation learning, the classifier is typically trained to mimic expert demonstrations. We generate demonstrations by sampling paths that lead to a local aesthetic maxima. Only demonstrations where the captured image exceeds the threshold for the current region are considered.

%is based on training a classifier to predict the current action to take from the current state to imitate the demonstrations provided by an expert, inspired by the methods used by \cite{jiang2020example} and \cite{huang2019learning} for autonomous cinematography. Since we don't have ground truth trajectories, we construct a training dataset by densely sampling paths that lead to the closest local maxima in terms of aesthetic score. For each path, we only add it to the training set if the local maxima achieves an aesthetic score that exceeds the aesthetic threshold for the current region.

\paragraph{Key Frame Selection} explores the scene, and backtracks to the most aesthetic view seen. This is reminiscent of key frame selection for video summarization \cite{money2008video, apostolidis2021video}. 
%and is inspired by related autonomous cinematography works \cite{cinematography}. In cinematography, the objective is typically associated with tracking an object or actor.
Since we do not have a predefined object of interest to track and scenes may be static, we cannot apply cinematography work \cite{jiang2020example, bonatti2020autonomous, huang2019blearning, huang2019learning} to generate trajectories. The agent instead explores the environment uniformly.

\paragraph{Greedy} selects an action at each position by executing every possible movement action, undoing that action, and finally selecting the action that improves the aesthetic score the most. If all movement actions would reduce the aesthetic score, CAPTURE is selected. Due to the large number of actions that the policy needs for a single effective step, this policy is very inefficient, but can eventually reach a local maxima if enough steps are executed. 
%is a policy that takes every movement and turning action at every single time step, and then takes the action that improves the aesthetic score the most. If no such action exists, then the policy takes the CAPTURE action and terminate. Due to the large number of actions that the policy needs to take for a single step, this baseline can be very inefficient but can eventually reach a local maxima if enough steps are taken.

\begin{figure}[ht!]
     \centering
     \includegraphics[width=\linewidth]{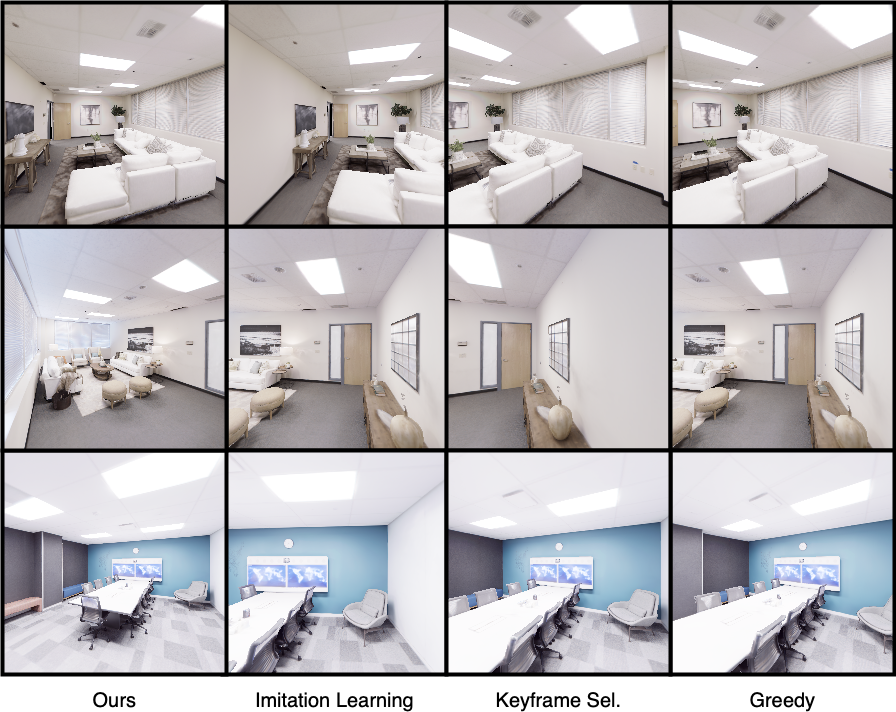} 
    \caption{\textbf{Simulation Visualizations.} We visualize photos captured by our method AutoPhoto against the strongest baselines on the Replica dataset. Note that AutoPhoto tends to better frame furniture compared to other methods.}
    \vspace{-4mm}
    \label{fig:sim_results}
\end{figure}

\begin{table}[!ht]
    % \small
    \centering
        \caption{\textbf{Accuracy in Achieving the Aesthetic Threshold.} We show the percentage of photos selected by each policy that achieve the aesthetic threshold. Refer to main text for descriptions of the baseline policies. ($\pm \sigma_{stderr}$)}
    \begin{tabularx}{0.8\linewidth}{|c|c|c|}%|c|c|} 
    \hline

    \makecell[c]{\emph{Metric}} & \multicolumn{2}{c|}{\makecell[c]{$\phi(s_T) > \tau_{aes}$ (\%)}} \\ % & \multicolumn{2}{c|}{\makecell[c]{Final Image  $>$ Initial Image  (\%)}} \\
    \hline
      \emph{Dataset} & \makecell[c]{Gibson} & Replica  \\ %& \hspace{7mm} Gibson \hspace{7mm} & Matterport
    \hline
    \makecell[l]{Random} & 13.3 $\pm$ 0.8 &  14.3 $\pm$ 0.8 \\ %& 32 &  \\ 
    \makecell[l]{Rule of Thirds \cite{visigrapp, auto_composition}} & 19.1 $\pm$ 0.9 & 17.3 $\pm$ 0.9 \\
    \makecell[l]{Greedy} & 56.6 $\pm$ 1.1 & 56.2 $\pm$ 1.2\\ 
     \makecell[l]{Key Frame Sel. \cite{money2008video, apostolidis2021video}} & 56.8 $\pm$ 1.1 &  56.6 $\pm$ 1.2 \\ %& 55 & 44.2
    \makecell[l]{Imitation Learning \cite{huang2019learning}} & 57.8 $\pm$ 1.2 & 51.9 $\pm$ 1.1 \\
    \makecell[l]{\bf Our Method} & \textbf{81.7 $\pm$ 0.9} &  \textbf{77.8 $\pm$ 1.0} \\ %& $\sim$83 & 79.4 \\ 
    \hline
    \end{tabularx}
    \label{table:eval_threshold}
\end{table}

\subsection{Evaluation in Simulation \label{sec:eval_sim}}

To ensure that the model can generalize to unseen scenes, we evaluate performance on 20 unseen scenes from Gibson \cite{gibson}. We also evaluate on 18 scenes from Replica \cite{replica} to measure generalization to a different dataset. 
We show the quantitative evaluation results on Gibson and Replica in Table \ref{table:eval_threshold}. Our model performs significantly better than the baselines. The low performance of the Rule of Thirds policy suggests that careful selection of objects of interest by human experts instead of automatic selection through saliency is important. Further, the rule of thirds does not fully model human aesthetics preferences (which are better captured by the aesthetics model).
%The low performance of the rule of thirds policy is likely attributed to the fact that no pre-defined object of interest exists. 
%Careful selection of objects to highlight is important, which shows why using a learned aesthetic estimator to assess the photo quality is useful. 
While both Key Frame Selection and Greedy can achieve strong performance given unlimited time, their performance is limited by inefficient exploration of the environment. The accuracy achieved by these methods is below that of our method when the number of steps taken is set to be similar as our method is more efficient in finding aesthetic views. Note that the performance of our method on Replica is close to the performance on Gibson, indicating that our model generalizes well to views from a different dataset. In Fig. \ref{fig:sim_results} we show some qualitative results against the Key Frame Selection and Greedy baselines on the Replica dataset. AutoPhoto tends to take well-composed photos of furniture and other appropriate objects in the photo.

\subsection{Deployment in Real Life \label{sec:real_life}}

% \hubert{NOTE: Since we do not specify collision detection etc in our pipeline, we should say in this section: "In our experiments, the robot takes no action if the selected action would result in collision (for example, if the agent would move into an obstacle). The model is queried again until a valid action is given. Since the model contains a memory module, it is able to learn that repeated actions with no change in state means another action should be selected. Automatic collision detection can be implemented via onboard depth sensors." } \hadi{do we want to explicitly say that we need to manually intervene to prevent the collision or that the detection is not autonomous? Can say: After every action, the RL model is queries for the next action to run on the Jackal. If the proposed action would lead to a collision, then the RL model is queried again until a valid action is given.}

\begin{figure}[!t]
     \centering
     \includegraphics[width=\linewidth]{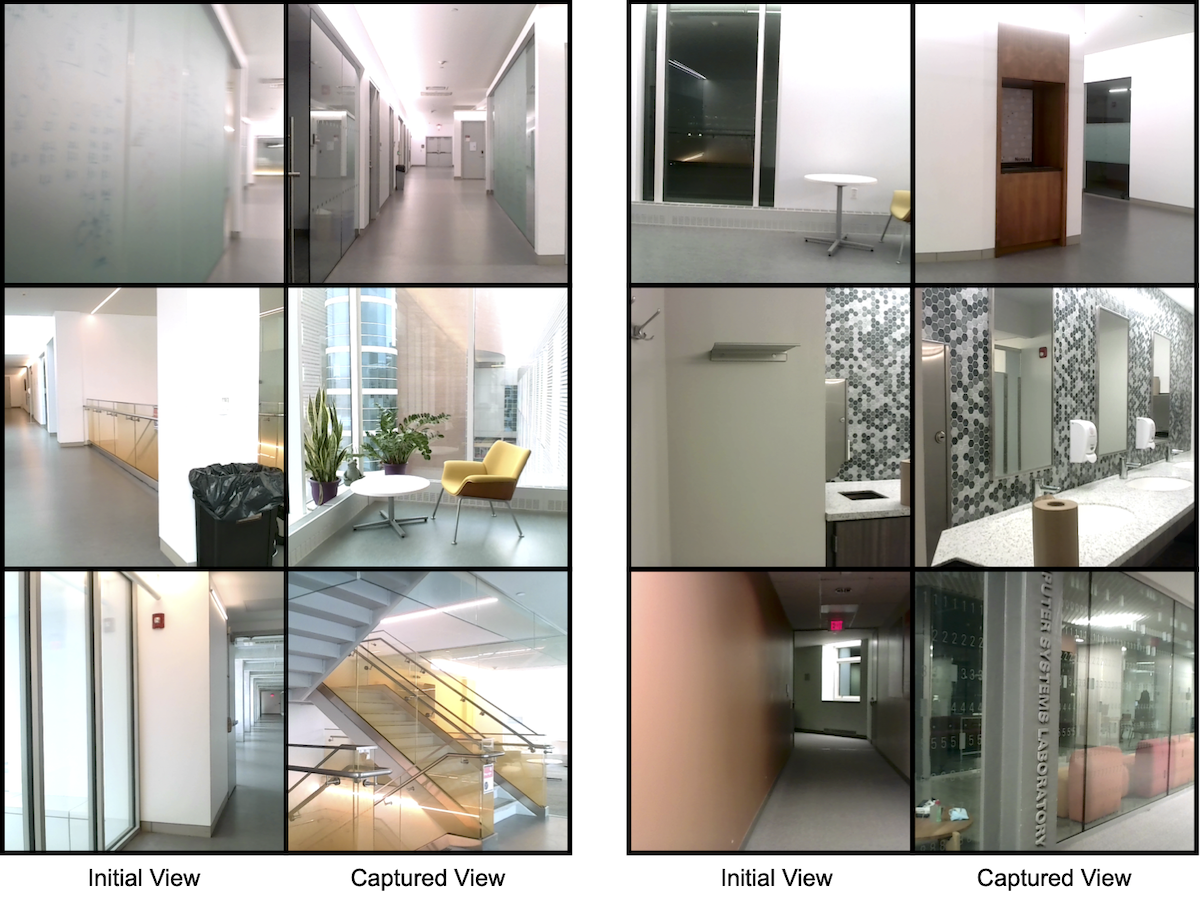}
    \caption{\textbf{Real World Visualizations.} AutoPhoto transfers from simulation to real life to capture well-composed photos.}
    \vspace{-2mm}
    \label{fig:real_results}
\end{figure}

We deployed our system in real-world settings on a Clearpath Jackal UGV (shown in Fig. \ref{fig:teaser}),
and used it to collect 64  photos of indoors environments. We attached a webcam to the Jackal at 1.5m above ground to approximately match the camera angle that would be typically used by human photographers. The input image is fed to the RL model to decide on which action to take, and then the command is sent to the robot through ROS. If the suggested action would lead to a collision, then the RL model is queried again until the proposed action is valid. Since the model contains a memory module, it is able to learn that repeated actions with no change in state means another action should be selected. In Fig. \ref{fig:real_results} we show sample photos that illustrates the initial views for the robot and the photos it captured.

Since it is not feasible to
densely sample photos in real-life to estimate an
aesthetic threshold, we conducted a user study on Amazon Mechanical Turk (AMT) to measure how often humans prefer the photos taken by AutoPhoto over the initial view of the environment.
Users are shown pairs of images, where each pair consists of an initial view and the final view captured by AutoPhoto. To reduce bias, both the order of pairs as well as images within each pair are shuffled. Users are asked to select which image is more aesthetic, or to select a ``tie'' option if both images are equally preferred. We select high quality workers through AMT's qualification system by only releasing the study to workers with a Master Qualification, over 95\% approval rate, and over 1000 previously approved tasks. For additional quality control, we included three sentinel pairs where the images within each pair are identical to each other. We only keep results from workers who correctly select ``tie'' for these sentinel pairs, and spend more than 1 second (median) per judgement across all pairs.
% photos that AutoPhoto captured and the initial view in randomized order, and ask the working to decide for each pair if they like the left or right image, or they consider them equally aesthetic. For quality assurance, we add sentinels like showing the same image both left and right, and duplicating some pairs more than once in the task. We expected users with good performance to choose the tie option when the pair is two copies of the same image, and that the user would make a consistent choice when presented when a pair more than once. We also check for the response time for each question to make sure that it is reasonable with respect to the human reaction time. Out of 46 users, we had 20 users pass the sentinels. We process the preferences by assigning the following scores for each pair: 1 if the user preferred image by AutoPhoto, 0.5 if they thought the aesthetic of the two images in the pair to be equivalent, and 0 otherwise.

We compute a preference score as: 1 if the user preferred the AutoPhoto image, 0 if the user preferred the initial image, and 0.5 if both images are equally preferred. Aggregating scores across 1792 judgements (28 valid user surveys), the mean preference score is $0.63 \pm 0.01$ (standard error). A t-test with null hypothesis being $0.5$ rejects the null hypothesis with $p < 0.05$, indicating that humans prefer the images taken by AutoPhoto.

\subsection{Ablation Studies \label{sec:ablation}}

We ran ablation studies for (a) the reward function and (b) the model architecture and (c) model actions. For the reward function, the non-terminal terms are ablated (Eq. \ref{eq:non-terminal-reward}). Specifically, the effect of the exploration term $\Gamma(\zeta)$ and the aesthetic score differences term $(\phi(s_{t+1}) - \phi(s_t))$ are measured.
%and finally remove all the terms except for the standard time penalty $\beta t$ and keep the final reward for the CAPTURE action. \\
For the architecture, we verify the importance of including LSTM memory in the model, and compare the performance multilayer input features for the agent versus only features from the last layer of the aesthetic model. For the model actions, we verify our decision to include fine and large rotation actions. In Table \ref{table:ablation} we show
the results of the ablations on Gibson and Replica. We observe that the exploration and score difference terms are equally important for performance, and removing them both causes the performance to degrade further. Regarding the architecture, we note that removing either memory or multilayer feature extraction lowers performance. Without an LSTM layer, actions cannot depend on the past, making scene exploration to assess the aesthetics of the local region impossible. %and decide what is considered aesthetic relative to the scene. 
With regards to input features, using multilayer features provides the agent with both high level and low level information encoded by the aesthetic model, leading to a boost in performance compared to only using the features extracted from the last layer. Finally, the addition of fine rotations (10 $^\circ$) and large rotations (90$^\circ$) allows the agent to better adjust view composition and explore the environment.

% Also note that the median number of steps of the
% policy without the LSTM layer is significantly higher on Matterport3D,
% which supports our claim that without a memory cell, the policy might
% not be able to terminate after a reasonable number of steps due to the
% partial observability of the problem setting. 

\begin{table}[!t]
    \small
    \centering
    \caption{\textbf{Ablation Results}. Our reward function terms, architecture design, and additional rotation actions are important for good performance.
    ($\pm \sigma_{stderr}$)
    %For the reward, both score difference and exploration terms play important roles. For our architecture, both LSTM memory as well as multilayer feature extraction are critical for good performance. ($\pm \sigma_{stderr}$)
    }
    %\hubert{Update table with std error} \hubert{include 30 deg turns only} \hadi{need to make it fit}}

    %Our method outperforms the variation with the modified reward functions, showing that the terms we include in the reward functions are critical for performance. We also observe that our design including LSTM memory module and using features extracted from multiple layers instead of relying on the features of the last layer improves performance significantly.
     % Also the results
    % show that removing the LSTM layer causes the model to take significantly larger number of steps on Matterport, supporting our claim that a memory cell is important to not fall in a loop when exploring the scene.
    
    \begin{tabularx}{\linewidth}{|c|c|c|c|}

    % \multicolumn{4}{c}{Rewards} \\
    \hline
    \makecell[c]{\multirow{2}{2.5em}{}} & \makecell[c]{\emph{Metric}} & \multicolumn{2}{c|}{\makecell[c]{  $\phi(s_T)> \tau_{aes}$ (\%)}} \\ % & \multicolumn{2}{c|}{\makecell[c]{Median \# steps}} \\
    \cline{2-4}
     & \emph{Dataset} & \makecell[c]{Gibson} & \makecell[c]{Replica} \\ %  & \makecell[c]{Gib.}  & \makecell[c]{Mat.} \\
    \hline
     %\makecell[l]{Term. Reward Only \& time penlaty} & 71 & 61 \\
    \multirow{3}{2.5em}{\!\!Reward} & \makecell[l]{w/o Score Diff. , Explor.} & 69.3 $\pm$ 1.0 & 58.4 $\pm$ 1.2 \\
    & \makecell[l]{w/o Score Diff.}  & 74.4 $\pm$ 1.0 & 66.7 $\pm$ 1.1  \\ 
   & \makecell[l]{w/o Explor.} & 72.1 $\pm$ 1.0 & 67.7 $\pm$ 1.1    \\ 
   \hline
    \makecell[c]{\multirow{2}{2.5em}{\!\!Architec.}} &  \makecell[l]{w/o LSTM} & 62.7 $\pm$ 1.1 & 55.4 $\pm$ 1.2 \\
    &    \makecell[l]{w/o Multilayer Feats.} & 62.3 $\pm$ 1.1 & 62.1 $\pm$  1.1\\
    
    \hline
    \makecell[c]{\multirow{1}{2.5em}{\!\!Actions}} &  \makecell[l]{w/o 10$^\circ$, 90$^\circ$ Rotation} & 73.7 $\pm$ 1.0 & 72.8 $\pm$ 1.0 \\
    % & \makecell[l]{\bf Our Method}  & \textbf{82} & \textbf{78}  \\
    \hline 
    & \makecell[l]{\bf Our Full Method} & \textbf{81.7 $\pm$ 0.9} &  \textbf{77.8 $\pm$ 1.0}  \\ 
    \hline
    % \multicolumn{3}{c}{Architecture} \\
    % \hline
    % \makecell[c]{\emph{Metric}} & \multicolumn{2}{c|}{\makecell[c]{  $\phi(s_T)> \tau_{aes}$ (\%)}} \\ % & \multicolumn{2}{c|}{\makecell[c]{Median \# steps}} \\
    % \hline
    % \emph{Dataset} & \makecell[c]{Gibson} & \makecell[c]{Replica} \\ %  & \makecell[c]{Gib.}  & \makecell[c]{Mat.} \\
    % \hline
    % \makecell[l]{w/o LSTM} & 62 & 57 \\
    %     \makecell[l]{w/o Multilayer Feats.} & 62 & 60 \\

    % \makecell[l]{\bf Our Method}  & \textbf{82} & \textbf{78}  \\
    % \hline
    \end{tabularx}
    \label{table:ablation}
    \vspace{-2mm}
\end{table}

%% file: sections/discussion.tex
\section{Conclusions}
In this work, we formulate the problem of photography as a POMDP and train an RL model to automatically capture aesthetically pleasing photos. We demonstrate our AutoPhoto system captures aesthetic photos in unseen scenes across simulation and real life. 
While our approach can generalize to domains beyond indoor scenes, it depends on having high quality simulated environments which could be harder to get for some domains than others. It is also important to select the aesthetic function appropriately as it influences the system behavior significantly.
To extend our work, a possible future direction is to expand the action space to a full 6 degrees of freedom on an aerial drone. Another interesting future direction is to include a user in the loop to ensure that photos taken capture the user intent.